\documentclass[11pt]{article}
\title{Asymptotics for The $k$-means}
\author{Tonglin Zhang \footnote{Department of Statistics, Purdue University, 250 North University Street,West Lafayette, IN 47907-2066, Email: tlzhang@purdue.edu}}
\usepackage{mathrsfs}
\usepackage{amsfonts}
\usepackage{amsmath}
\usepackage{wrapfig,lipsum,booktabs}
\usepackage{bm}
\usepackage{algorithm}
\usepackage{algpseudocode}
\usepackage{pifont}
\usepackage{epsfig}
\def\qed{\hfill$\diamondsuit$}
\newtheorem{defn}{Definition}

\newtheorem{thm}{Theorem}
\newtheorem{cor}{Corollary}
\newtheorem{lem}{Lemma}
\begin{document}
\def\qed{\hfill$\diamondsuit$}
\def\ed{\end{document}}
\def\SSB{S\negthinspace S\negthinspace B}
\def\MSB{M\negthinspace S\negthinspace B}
\def\SSW{S\negthinspace S\negthinspace W}
\def\MSW{M\negthinspace S\negthinspace W}
\maketitle
\def\eqalign#1{\null\,\vcenter{\openup\jot\ialign
              {\strut\hfil$\displaystyle{##}$&$\displaystyle{{}##}$
               \hfil\crcr#1\crcr}}\,}

\begin{abstract}

The $k$-means is one of the most important unsupervised learning techniques in statistics and computer science. The goal is to partition a data set into many clusters, such that observations within clusters are the most homogeneous and observations between clusters are the most heterogeneous. Although it is well known, the investigation of the asymptotic properties is far behind, leading to difficulties in developing more precise $k$-means methods in practice. To address this issue, a new concept called clustering consistency is proposed. Fundamentally, the proposed clustering consistency is more appropriate than the previous criterion consistency for the clustering methods. Using this concept, a new $k$-means method is proposed. It is found that the proposed $k$-means method has lower clustering error rates and is more robust to small clusters and outliers than existing $k$-means methods. When $k$ is unknown, using the Gap statistics, the proposed method can also identify the number of clusters. This is rarely achieved by existing $k$-means methods adopted by many software packages. 
\end{abstract}



\section{Introduction}
\label{sec:introduction}
Clustering is one of the most important unsupervised learning techniques for understanding the underlying data structures. The goal is to partition a data set into many subsets, called clusters, such that the observations within the subsets are the most homogeneous and the observations between the subsets are the most heterogeneous. Clustering is usually carried out by specifying a similarity or dissimilarity measure between observations. Examples include the $k$-means~\cite{hartigan1975,hartigan1979,macqueen1967,steinhaus1957}, the $k$-medians~\cite{cardot2012}, the $k$-modes~\cite{chaturvedi2001}, and the generalized $k$-means~\cite{bock2008,maranzana1963,zhanglin2021}, as well as many of their modifications~\cite{huang2005,jing2007,witten2010}. Among those, the $k$-means has been considered as one of the most straightforward and popular methods since it was proposed sixty years ago~\cite{jain2010,steinley2006}. Although it is well known, the investigation of the theoretical properties is still far behind, leading to difficulties in developing more precise $k$-means methods in practice. The goal of the present research is to propose a new concept called clustering consistency for the asymptotics of the $k$-means with a resulting clustering method better than the existing $k$-means methods adopted by many software packages, including those adopted by \textsf{R} and \textsf{Python}.

The research has three main contributions. The first is the development of clustering consistency, which is more appropriate than the previous criterion consistency used for decades for asymptotics of clustering methods. In the literature, theoretical properties of the $k$-means are investigated  by the performance of the sum-of-squares (SSQ) criterion function, which is classified as criterion consistency by the article. The difference is that clustering consistency focuses on the asymptotic properties of the partition resulted by the $k$-means but not criterion consistency focuses on the asymptotic properties of the SSQ criterion used by the $k$-means. The second is the derivation of a better $k$-means method using clustering consistency. It is better because any local solution of the proposed $k$-means method is clustering consistent, implying that we successfully avoid the NP-hardness issue~\cite{mahajan2012} for the global solution of the $k$-means problem. This property is violated if other $k$-means methods are used. The third is estimation of the number of clusters when $k$ is unknown. This is a major concern in practice. Our method can be used to estimate the number of clusters if it is combined with the well known Gap statistics~\cite{tibshirani2001}. 

Among the large amount of the literature on the categories of unsupervised clustering learning, the $k$-means is considered as the simplest and also stands out to be the most popular one. It belongs to the category of partitioning clustering, which can be interpreted by the centroidal Voronoi tessellation method in mathematics~\cite{du2002}. Besides the $k$-means, partitioning clustering also includes the $k$-medians, the $k$-modes, and the generalized $k$-means. Other clustering categories proposed in the literature include hierarchical clustering~\cite{zhao2005}, fuzzy clustering~\cite{trauwaert1991}, density-based clustering~\cite{ester1996,kriegel2011}, and model-based clustering~\cite{fraley2002,lau2007}. 

The $k$-means is an iterative method that tries to partition a data set into $k$ distinct non-overlapping and non-empty clusters. In particular, let ${\cal D}=\{{\bm x}_1,\dots,{\bm x}_n\in\mathbb{R}^p\}$ be the data set composed by $n$ observations and $\mathscr{C}_k=\{C_1,\dots,C_k: \bigcup_{r=1}^k C_r={\cal D}, C_r\not=\emptyset, C_{r}\cap C_{r'}=\emptyset, r\not=r' \}$ be a size-$k$ partition of ${\cal D}$. The $k$-means estimates the partition by 
\begin{equation}
\label{eq:k-means criterion}
\widehat{\mathscr{C}_k}=\{\hat C_1,\dots,\hat C_k\}=\mathop{\arg\!\min}_{\mathscr{C}_k}SSQ(\mathscr{C}_k),
\end{equation}
where 
\begin{equation}
\label{eq:SSQ criterion in k-means}
SSQ(\mathscr{C}_k)=\sum_{r=1}^k\sum_{i\in{C_r}}\|{\bm x}_i-{\bm c}_r\|^2,
\end{equation}
is the SSQ criterion and ${\bm c}_r\in\mathbb{R}^p$ is the centroid of $C_r$. The $k$ centroids ${\bm c}_1,\dots,{\bm c}_k$ are treated as unknown parameters.  In practice, $\widehat{\mathscr{C}_k}$ and $\hat{\bm c}_1,\dots,\hat{\bm c}_k$ (i.e., the corresponding estimates of the $k$ centroids) are computed iteratively. A $k$-means method starts with initial guesses of $\hat{\bm c}_1,\dots,\hat{\bm c}_k$. They are used to compute the optimal partition of the data set.  The centroids are updated after that. By iterations, the final estimates of the partition and the centroids are derived. It has been pointed out by many researchers that the final answer may not be the global solution because the output depends on the initial guesses of the centroids. To avoid a local optimizer, many researchers recommend performing the algorithm multiple times with different initial guesses of the centroids~\cite{falkenauer2001,hartigan1975}. It has been pointed out that this may induce thousands of replications in searching the global optimizer even when $n$ is a few thousand~\cite{steinley2003}. 

The theoretical properties of $k$-means algorithms highly depend on the initialization of the centroids. The global solution of~\eqref{eq:k-means criterion} is NP-hard~\cite{mahajan2012}. To avoid this issue, many researchers investigate local solutions. The performance of the basic (i.e., the earliest) $k$-means algorithm is bad because it uses randomly generated initial centroids. To overcome the difficulty, methods using initial centroids far away from each other have been proposed. Among those, the $k$-means++~\cite{arthur2007}, which is the default of \textsf{python}, is considered as one of the best methods. Although it is not the default of \textsf{R}, the $k$-means++ has been incorporated by library \textsf{LICORS}.  Although the performance of the $k$-means++ is better than that of the basic $k$-means, we identify a better $k$-means method using clustering consistency.

The article is organized as follows. In Section~\ref{sec:related work}, we review previous criterion consistency for the $k$-means. In Section~\ref{sec:main result}, we propose clustering consistency for the $k$-means and use it to develop a new $k$-means method under the $k$-means model. In Section~\ref{sec:extension}, we extend our findings beyond the $k$-means model. In Section~\ref{sec:simulation}, we evaluate the performance of our method with the comparison to the previous $k$-means methods by Monte Carlo simulations. In Section~\ref{sec:application}, we apply our method to a real world data set. In Section~\ref{sec:discussion}, we provide a discussion. We put all of the proofs in the Appendix.

\section{Related Work}
\label{sec:related work}

In the literature, asymptotics for the $k$-means is evaluated by the properties of the SSQ-criterion. It is classified as criterion consistency in the article. Rather than the proposed clustering consistency which focuses on the local solution of the $k$-means, the previous criterion consistency focuses on the the global solutions only. The interest is to show the final $SSQ(\widehat{\mathscr{C}_k})$ (assume that it is the global solution) satisfies
\begin{equation}
\label{eq:SSQ the true partition}
{1\over n}[SSQ(\widehat{\mathscr{C}_k})-SSQ(\mathscr{C}_{k,T})]\stackrel{P}\rightarrow 0, {\rm as}\ n\rightarrow\infty,
\end{equation}
where $SSQ(\mathscr{C}_{k,T})=\sum_{r=1}^k\sum_{i\in{C_{r,T}}}\|{\bm x}_i-{\bm\mu}_r\|^2$, ${\bm\mu}_r$ is the true center of the $r$th cluster, and $\mathscr{C}_{k.T}$ is the true partition. 

Based on criterion consistency,  as early as in 1967, MacQueen~\cite{macqueen1967} obtained weak consistency for the $k$-means. He showed that the corresponding SSQ criterion converges almost surely, but he did not prove consistency of the cluster centroids. In 1978, Hartigan~\cite{hartigan1978} obtained consistency and asymptotic normality of the SSQ criterion in the one-dimensional case when the observations are splitted into two clusters. In 1981, Pollard~\cite{pollard1982a} provided conditions for almost sure convergence of the cluster centers as the sample size increases to infinity in the high-dimensional case. In 1984, Wong~\cite{wong1984} obtained consistency of the univariate $k$-means when $k$ goes to infinity with $n$. In 2015, Terada~\cite{terada2015} provided strong consistency of the SSQ criterion when the principal component analysis (PCA) approach is applied to the $k$-means for dimension reduction. In 2022, Chakrabary and Das~\cite{chakrabarty2022} established strong consistency of estimators of cluster centroids. In all of these, the interest is consistency of the SSQ-criterion and the estimators of the centroids but not the estimator of the partition itself.

To implement~\eqref{eq:SSQ the true partition} for criterion consistency, it is necessary to provide a nice definition for {\it true clusters} and {\it true clustering}. It is well known that there is no agreed definition of what the true clustering is if only the observed data set ${\cal D}$ is used. To interpret this concept, a previous method treats the global solution of~\eqref{eq:k-means criterion} as the true clustering, but this is considered as an inappropriate approach~\cite{henning2015}. In the situation of clustering analysis, it is assumed that the true cluster labels are unknown. The goal of a clustering method is to assign labels to all ${\bm x}_i$. Without using the true labels, it is hard to define true partition and true clustering. Therefore, we assume that the true cluster labels are disclosed in the definition of $\mathscr{C}_{k,T}$ although they cannot be used to formulate a usable clustering method, meaning that we treat true clustering as an unusable clustering method. Using this idea, we obtain the formal definition of the true clustering and true partition. If this is adopted, then the criterion consistency given by~\eqref{eq:SSQ the true partition} is well defined. We provide interpretations of the previous  criterion consistency and the proposed clustering consistency. 

\section{Main Result}
\label{sec:main result}

We use a complete data model and an incomplete data model for the development of our method. In the complete data model, we assume that the true cluster labels are disclosed. Each ${\bm x}_i$ is associated with an indicator vector ${\bm z}_i=(Z_{i1},\dots,Z_{ik})^\top$ with $Z_{ir}\in\{1,0\}$ and $\sum_{r=1}^k Z_{ir}=1$ for its cluster assignment. The complete data model assums that both ${\bm x}_i$ and ${\bm z}_i$ are available, leading to the complete date set as ${\cal D}_c=\{({\bm x}_i:{\bm z}_i):i=1,\cdots,n\}$.  The incomplete data model assumes that the true cluster labels are not observed. It treats ${\bm z}_i$ as unobserved latent variables, leading to the incomplete data set as ${\cal D}=\{{\bm x}_i:i=1,\dots,n\}$. We use ${\cal D}_c$ to define the true clustering. Because a usable clustering method can only use ${\cal D}$, the true clustering is treated as an unusable method. The goal of clustering consistency is to show that the estimate of the partition resulted by a usable clustering method is asymptotically identical to the partition resulted by the true clustering method, indicates that it is asymptotically equivalent to the case when the cluster assignments are disclosed. Due to the NP-hardness issue of the global solution of the $k$-means problem, we provide a new $k$-means method with clustering consistency satisfied by its local solutions, implying that the NP-hardness issue is not a concern in our method. 

\subsection{Gaussian Mixture Models}
\label{eq:gaussian mixture models}

We evaluate the asymptotic properties of the general $k$-means method and develop our new $k$-means method under the framework of the Gaussian mixture models (GMMs). Theoretically, a GMM is a special case of mixture models. A mixture model is defined by a hierarchical modeling approach. The first level specifies the distributions of ${\bm z}_i$ for the ground truth. The second level specifies the distribution of the observations given ${\bm z}_i$. The mixture model can be interpreted by either a complete data model or an incomplete data model. The complete data model assumes that the complete data set ${\cal D}_c$ is available, leading to the underlying distribution as
\begin{equation}
\label{eq:underlying distribution of the data incomplete data}
{\bm x}_i|{\bm z}_i\sim^{iid} \sum_{r=1}^k Z_{ir}f_{r}({\bm x}_i)=\prod_{r=1}^k [f_r({\bm x}_i)]^{Z_{ir}}.
\end{equation}
and the loglikelihood function as
\begin{equation}
\label{eq:loglikelihood of the complicate data model}
{\bm\ell}_c({\bm\theta})=\sum_{i=1}^n\sum_{r=1}^k Z_{ir}\log[f_r({\bm x}_i)],
\end{equation}
where ${\bm\theta}$ the vector composed by all parameters involved in the model. 

The incomplete data model treats ${\bm z}_i$ as unobserved latent variables. It is assumed that only  ${\cal D}=\{{\bm x}_i:i=1,\dots,n\}$ is available. Suppose that ${\bm z}_i$ are iid from a Dirichlet  distribution with probability vector ${\bm\pi}=(\pi_1,\dots,\pi_k)^\top$. By integrating out the latent variables ${\bm z}_i$ from~\eqref{eq:underlying distribution of the data incomplete data}, the underlying distribution of the incomplete data model becomes ${\bm x}_i\sim^{iid} \sum_{r=1}^k \pi_r f_r({\bm x}_i)$.
A GMM is derived if $f_r({\bm x}_i)$ is specified the PDF of ${\cal N}({\bm\mu}_r,{\bm\Sigma}_r)$ as
\begin{equation}
\label{eq:PDF component general GMM}
f_r({\bm x}_i)=\varphi({\bm x}_i;{\bm\mu}_r,{\bm\Sigma}_r)=-{1\over (2\pi)^{p\over2}|{\det({\bm\Sigma}_r)|^{1\over2}}}e^{-{1\over 2}({\bm x}_i-{\bm\mu}_r)^\top{\bm\Sigma}_r^{-1}({\bm x}_i-{\bm\mu}_r)}
\end{equation}
 leading to underlying model in the incomplete data model as
\begin{equation}
\label{eq:underlying distribution of GMM}
{\bm x}_i\sim^{iid}\sum_{r=1}^k \pi_r {\cal N}({\bm\mu}_r,{\bm\Sigma}_r),
\end{equation}
where ${\bm\mu}_r$ is the mean vector (i.e., the true center) and ${\bm\Sigma}_r$ is the variance-covariance matrix of the $r$th cluster. 

The mixture probability (i.e.,  weight) $\pi_r$ is positive  and  satisfies $\sum_{r=1}^k \pi_r=1$. The GMM treats $\pi_r$, ${\bm\mu}_r$, and ${\bm\Sigma}_r$ as unknown parameters. By treating the latent variables as missing, the MLE of the parameters can be computed by the EM-algorithm~\cite{dempster1977}. If ${\bm\Sigma_r}=\sigma^2{\bf I}$ for all $r=1,\dots,k$, then the GMM becomes the $k$-means model. Because the EM algorithm is a method for the MLE when missing values are present~\cite{wu1983}, it can also be used to study the $k$-means problem. In this case, we need to assume that ${\bm\mu}_1,\dots,{\bm\mu}_k$ are all different in~\eqref{eq:underlying distribution of GMM} to be consistent with the $k$-means problem.

We briefly review the EM algorithm for the GMM given by~\eqref{eq:underlying distribution of the data incomplete data} with $f_r({\bm x}_i)$ specified by~\eqref{eq:PDF component general GMM}, which has bee formulated previously by many authors~\cite[e.g.]{nityasuddhi2003}. Let $\pi_r^{(t)}$, ${\bm\mu}_r^{(t)}$, and ${\bm\Sigma}_r^{(t)}$ be the $t$th iterative values of the MLEs of $\pi_r$, ${\bm\mu}_r$, and ${\bm\Sigma}_r$ (denoted by $\hat\pi_r$, $\hat{\bm\mu}_r$, and $\hat{\bm\Sigma}_r$, respectively) in the EM for the GMM defined by~\eqref{eq:underlying distribution of GMM}. The E-step of the EM-algorithm predicts ${\bm z}_i$ as 
\begin{equation}
\label{eq:E-step of EM}
 Z_{ir}^{(t+1)}={\pi_r^{(t)}\varphi({\bm x}_i|{\bm\mu}_r^{(t)},{\bm\Sigma}_r^{(t)})\over \sum_{j=1}^k \pi_j^{(t)}\varphi({\bm x}_i|{\bm\mu}_j^{(t)},{\bm\Sigma}_j^{(t)})} .
\end{equation}
The M-step of the EM-algorithm estimates the parameters as
\begin{equation}
\label{eq:M-step of EM}
\eqalign{
\pi_r^{(t+1)}=&{1\over n}\sum_{i=1}^n  Z_{ir}^{(t+1)},   \cr
{\bm\mu}_r^{(t+1)}=& {\sum_{i=1}^n Z_{it}^{(t+1)}{\bm x}_i\over \sum_{i=1}^n Z_{ir}^{(t+1)}}, \cr
{\bm\Sigma}_r^{(t+1)}=&  {\sum_{i=1}^n Z_{it}^{(t+1)}({\bm x}_i-{\bm\mu}_r^{(t+1)})^\top ({\bm x}_i-{\bm\mu}_r^{(t+1)})\over \sum_{i=1}^n Z_{ir}^{(t+1)}}.\cr
}
\end{equation}
In the end, the EM-algorithm estimates the partition by
\begin{equation}
\label{eq:the partition EM}
\hat C_{r}=\{i: \hat Z_{ir}=\mathop{\arg\!\max}_{j\in\{1,\dots,k\}} \hat Z_{ij}\},\ r=1,\dots,k,
\end{equation}
where $\hat{\bm z}_i$ is the $i$th final imputed vector of the ground truth given by iterating~\eqref{eq:E-step of EM} and~\eqref{eq:M-step of EM}. By taking ${\bm\Sigma}_r=\sigma^2{\bf I}$ in the GMM, we obtain the $k$-means model with
\begin{equation}
\label{eq:the distribution of the rth cluster}
f_r({\bm x}_i)=\varphi({\bm x}_i;{\bm\mu}_r,\sigma^2{\bf I})=-{1\over(2\pi)^{p\over 2}\sigma^{p}}e^{-{\|{\bm x}_i-{\bm\mu}_r\|^2\over 2\sigma^2}}
\end{equation}
in~\eqref{eq:underlying distribution of the data incomplete data}, leading to the EM-algorithm for the $k$-means model. 

The $k$-means method does not use the EM-algorithm. Instead, it directly computes the next iterative value of ${\bm z}_i$ given the previous ${\bm\mu}_1^{(t)},\dots,{\bm\mu}_k^{(t)}$ as
\begin{equation}
\label{eq:prediction of ground truth latent variables}
Z_{ir}^{(t+1)}=\left\{ \begin{array}{cc}   1,  &  r=\mathop{\arg\!\min}_{j\in\{1,\dots,k\}}\|{\bm x}_i-{\bm\mu}_j^{(t)}\| , \cr 0 , & {\rm otherwise}, \end{array}\right.     
\end{equation}
and the next iterative value of the MLE of ${\bm\mu}_r$ as
\begin{equation}
\label{eq:estimator of mu k-means}
{\bm\mu}_r^{(t+1)}={\sum_{i=1}^n Z_{ir}^{(t+1)}{\bm x}_i\over \sum_{i=1}^n Z_{ir}^{(t+1)}}.
\end{equation}
In the end, it estimates the partition by 
\begin{equation}
\label{eq:estimator of partition kmeans}
\hat C_r=\{i: \hat Z_{ir}=1\},\  r=1,\dots,k,
\end{equation}
where $\hat{\bm z}_i=(\hat Z_{i1},\dots,\hat Z_{ik})^\top$ is the final imputed vector of the ground truth given by iterating~\eqref{eq:prediction of ground truth latent variables} and~\eqref{eq:estimator of mu k-means}. Neither the EM-algorithm nor the $k$-means uses the ground truth ${\bm z}_i$ in estimating the clusters. Instead, they use the imputed $\hat{\bm z}_i$. Therefore, they are usable clustering methods. 

\subsection{Clustering Consistency}
\label{subsec:clustering consistency}

The usage of~\eqref{eq:the partition EM} by the EM-algorithm or~\eqref{eq:estimator of partition kmeans} by a $k$-means method indicates that we should use ${\bm z}_i$ to define the true partition and the true clustering. In the literature, the definition of the true partition and true clustering is considered as a concern in the study of theoretical properties of clustering methods. Several approaches have been proposed previously~\cite{henning2015}, but none of them use ${\bm z}_i$ to define the true partition or the true clustering. To overcome the difficulty, we provide the following definition.

\begin{defn}
\label{defn:definition of true clustering}
The true partition for a mixture model is $\mathscr{C}_{k,T}=\{C_{1,T},\dots,C_{k,T}\}$ with $C_{r,T}=\{i: Z_{ir}=1\}$ for $r=1,\dots,k$. A clustering method is said true clustering if it uses ${\bm z}_i$ to assign the labels of clusters. A clustering method is said usable if it does not use ${\bm z}_i$ to assign the clusters.  
\end{defn}

The true clustering does not make any mistakes in assigning the cluster labels. The true clustering given by Definition~\ref{defn:definition of true clustering} is not usable because it uses ${\bm z}_i$. It is generally impossible to formulate a clustering method which can always result $\mathscr{C}_{k,T}$ without using ${\bm z}_i$ when $n$ is finite. Therefore, it is important to compare the clusters assigned by a usable clustering method with those assigned by the true clustering method as $n\rightarrow\infty$, leading to the clustering consistency developed by the current research. 

\begin{defn}
\label{defn:definition of clustering consistency}
We say that a $k$-means method resulting $\widehat{\mathscr{C}_k}$ is clustering consistent, denoted by $\widehat{\mathscr{C}_k}\stackrel{P}\rightarrow\mathscr{C}_{k,T}$, if
\begin{equation}
\label{eq:partition converges in probability}
\lim_{n\rightarrow\infty}{\rm Pr}(\widehat{\mathscr{C}_k}=\mathscr{C}_{k,T})=1.
\end{equation}
We say that the $k$-means method is clustering inconsistent if~\eqref{eq:partition converges in probability} is violated.
\end{defn}

Clustering consistency can be extended to arbitrary (usable) clustering methods, because the only requirement is to ensure that~\eqref{eq:partition converges in probability} holds. Because the labels of clusters are not specified in the data, clustering consistency given by Definition~\ref{defn:definition of clustering consistency} does not mean $\hat C_{r}\stackrel{P}\rightarrow C_{r,T} $ for every $r$. Instead, it only needs that there is a permutation $\tau$ on $\{1,\dots,k\}$, such that $\hat C_{\tau(r)}\stackrel{P}\rightarrow C_{r,T}$ for every $r$.  Therefore, Definition~\ref{defn:definition of clustering consistency} contains $k!$ labeling cases. In the case when the ground truth variables are provided but not used in the $k$-means method, to check whether clustering consistency holds, we need to find $\tau$. This kind of problems can be solved by the well known Hungarin algorithm in mathematics. 

To obtain clustering consistency, we need to impose  regularity conditions for ${\bm\mu}_1,\dots,{\bm\mu}_k$ and $\sigma^2$. The following example shows that the minimum distance between ${\bm\mu}_1,\dots,{\bm\mu}_k$ should go to infinity as $n\rightarrow\infty$ if $\sigma^2$ is fixed.

{\it Example 1:} Suppose that $\sigma^2=1$, $k=2$, and $\pi_1=\pi_2=1/2$ in~\eqref{eq:the distribution of the rth cluster}. Then, the centroid of the first cluster is ${\bm c}_1={\rm E}[{\bm x}|\|{\bm x}-{\bm\mu}_1\|\le \| {\bm x}-{\bm\mu}_2\|] $ and the center of the second cluster is ${\bm c}_2={\rm E}[{\bm x}|\|{\bm x}-{\bm\mu}_1\|\ge \| {\bm x}-{\bm\mu}_2\|]$, which are  functions of ${\bm\mu}_1$, ${\bm\mu}_2$, and $\sigma^2$. Both ${\bm c}_1$ and ${\bm c}_2$ are on the line segment between ${\bm\mu}_1$ and ${\bm\mu}_2$. Let $\hat{\bm c}_1$ and $\hat{\bm c}_2$ be the estimates of ${\bm c}_1$ and ${\bm c}_2$ provided by the global solution. Then, $(\hat{\bm c}_1+\hat{\bm c}_2)/2\stackrel{P}\rightarrow({\bm\mu_1}+{\bm\mu}_2)/2$ as $n\rightarrow\infty$ by the asymptotic theory of M-estimation. Asymptotically, the $k$-means assigns ${\bm x}_i$ to the first cluster if $\|{\bm x}_i-{\bm\mu}_1\|\le \|{\bm x}_i-{\bm\mu}_2\|$ or the second cluster otherwise. Because ${\rm Pr}\{\|{\bm x}-{\bm\mu}_1\|\le \|{\bm x}-{\bm\mu}_2\||{\bm x}\in C_{2,T}\}={\rm Pr}\{\|{\bm x}-{\bm\mu}_2\|\le \|{\bm x}-{\bm\mu}_1\||{\bm x}\in C_{1,T}\}$ is positively bounded from $0$ asymptotically,  we do not have $\hat C_{\tau(1)}\stackrel{P}\rightarrow C_{1,T}$ and $\hat C_{\tau(2)}\stackrel{P}\rightarrow C_{2,T}$ as $n\rightarrow\infty$, implying that any $k$-means methods cannot be clustering consistent  under this setting. 

The main issue in Example 1 is that $\|{\bm\mu}_1-{\bm\mu}_2\|$ is bounded when $\sigma^2$ is fixed as $n\rightarrow\infty$, leading to clustering inconsistency of all possible $k$-means methods. When $n$ becomes large, the size of the first cluster grows. Sooner or later, it will cover ${\bm\mu}_2$. This also happens in the second cluster. Therefore, to obtain clustering consistency, we need to assume that
\begin{equation}
\label{eq:minium distance}
d=\min\{ \|{\bm\mu}_r-{\bm \mu}_{r'}\|: r\not=r'\}\rightarrow\infty
\end{equation}
as $n\rightarrow\infty$. This has also been pointed out in the case when the $k$-means is applied to group functional data in two clusters  (i.e., $k=2$) by a concept called asymptotic perfectness~\cite{delaigle2019}. The difference is that asymptotic perfectness focuses on the global solution but clustering consistency also investigates the local solutions. Using clustering consistency, we conclude if the speed for $d$ to approach $\infty$ is faster than that of $\log^{1/2}{n}$ (when $\sigma^2$ is fixed), then any location solution provided by the proposed $k$-means method (i.e., given by Algorithm~\ref{alg:improved k-means algorithm}) is clustering consistent. This property may not be satisfied if other $k$-means methods are used. Thus, we can use local solutions of our method. If other $k$-means method are used, then we should use the global solution. Then, we have the following theorems.

\begin{thm}
\label{thm:clustering consistency of the k means given k}
If (a) ${\bm x}_1,\dots,{\bm x}_n$ are independently collected from~\eqref{eq:underlying distribution of GMM} with ${\bm\Sigma}_r=\sigma^2{\bf I}$, (b) the initial centroids are provided by data points in different clusters, (c) $\mathop{\lim\!\inf}_{n\rightarrow\infty} d/[4\sigma(2\log{n})^{1/2}]>1$, and (d) $\lim_{n\rightarrow\infty} p\log[d^2/(p\sigma^2)]/\log{n}=0$, then the $k$-means method is clustering consistent.
\end{thm}

 Theorem~\ref{thm:clustering consistency of the k means given k} can be implemented even if $\sigma^2$ and $p$ vary with $n$. Condition (d) is not needed if $p$ is a constant. Therefore, Condition (c) is more important. It has two special cases. The first case assumes that $\sigma^2$ is a constant. We need that $d$ grows to infinity with $n$. The second assumes that $d$ is a constant. We need that $\sigma^2$ goes to $0$ as $n$ goes to infinity. We have the following Corollary. 

\begin{cor}
\label{cor:clustering consistent sphere constant sigmasq and p}
Assume that Conditions (a) and (b) of Theorem~\ref{thm:clustering consistency of the k means given k} are satisfied, and $p$ is a constant. If (e) (i) $\sigma^2$ is a constant and $\lim\!\inf_{n\rightarrow\infty} d/(\log{n})^{1/2}>4\sqrt{2}\sigma$ or (ii) $d$ is a constant and $\lim\!\sup_{n\rightarrow\infty} \sigma^2\log{n}< d^2/32$, then the $k$-means method is clustering consistent.
\end{cor}

When $p$ and $\sigma^2$ does not vary with $n$,  we can obtain a $k$-mean method with clustering consistency if $d\rightarrow\infty$ with speed faster than $\log^{1/2}{n}$. To ensure clustering consistency, we also need to set the initial cluster centroids in distinct clusters. This is usually violated in many well known $k$-means methods. For instance, the \textsf{kmeans} function of \textsf{R} selects the initial centroids by randomness, which cannot guarantee that they belong to distinct clusters (as $n\rightarrow\infty$). The \textsf{kmeanspp} of~\textsf{R} and the \textsf{kmeans} function of \textsf{python} selects the next centroid with a probability inversely proportional to the squares of the minimum distance to the previous selected centroids, which cannot guarantee that they belong to distinct clusters either. Therefore, it is more important to propose a new $k$-means method satisfying  Condition (b) of Theorem ~\ref{thm:clustering consistency of the k means given k}.

\subsection{A New $k$-means Method}
\label{subsec:a new k-means method}

We devise an approach to select the $k$ initial centroids by Condition (b) of Theorem~\ref{thm:clustering consistency of the k means given k}. Our approach contains $k$ steps. Each step provides an initial centroid. In the end, it provides the $k$ initial centroids. In the first step, we randomly select the first initial centroid $\tilde{\bm c}_1$. In the second step, we select the second initial centroid $\tilde{\bm c}_2$ as the data point with the maximum distance to the first initial centroid. Therefore, $\tilde{\bf c}_2$ satisfies
\begin{equation}
\label{eq:condition for the second initial centroid}
\tilde{\bf c}_2=\mathop{\arg\!\max}_{{\bm x}\in{\cal D},{\bm x}\not=\tilde{\bm c}_1} \|{\bm x}-\tilde {\bm c}_1\|.
\end{equation}
 From the third step (i.e., the $r$th step for $r\ge 3$), we define the distance measure
\begin{equation}
\label{eq:distance to the previous selected centroids}
D({\bm x})=\min_{\tilde{\bm c}\in\tilde C_{prev} }(\|{\bm x}-\tilde{\bm c}\|),
\end{equation}
 where $\tilde {\cal C}_{prev}=\{\tilde{\bm c}_1,\dots,\tilde{\bm c}_{r-1}\}$ is the set of the initial centroids selected by the previous steps. We select the $r$th initial centroid by
\begin{equation}
\label{eq:the current initial centroid}
\tilde{\bf c}_r=\mathop{\arg\!\max}_{{\bm x}_i\in{\cal D}\backslash\tilde C_{prev}} D({\bm x}_i).
\end{equation}
Therefore, $\tilde{\bm c}_r$ satisfies the max-min principal. It is the data point which maximizes the minimum distance to the previous selected initial centroids. In the end, we obtain the set of $k$ initial centroids, denoted by $\tilde{\cal C}=\{\tilde{\bm c}_1,\dots,\tilde{\bm c}_k\}$, leading to the proposed $k$-means method by Algorithm~\ref{alg:improved k-means algorithm}.

\begin{algorithm}
\caption{\label{alg:improved k-means algorithm}Algorithm for the improved $k$-means clustering}
\begin{algorithmic}[1]
\Statex{{\bf Input}: Data set ${\cal D}=\{{\bm x}_1,\dots,{\bm x}_n\}$ and the number of clusters $k$}
\Statex{{\bf Output}:  labels $1,\dots,k$ for each ${\bm x}_i\in{\cal D}$.}
\Statex{\it Initialization}
\State{Choose the first initial centroid $\tilde{\bm c}_1$ from ${\cal D}$ randomly}
\State{For each ${\bm x}_i\in{\cal D}$ distinct from $\tilde{\bm c}_1$, compute $\|{\bm x}_i-\tilde{\bm c}_1\|$. Choose the second initial centroid $\tilde{\bm c}_2$ as the data point with the maximum  $\|{\bm x}_i-\tilde{\bm c}_1\|$ value.}
\State{Starting from the third initial centroid, compute $D({\bm x}_i)$ by~\eqref{eq:distance to the previous selected centroids} for all ${\bm x}_i\in{\cal D}$  but distinct from the previous selected initial centroids.  Choose the current initial centroid as the data point with the maximum $D({\bm x}_i)$ value. Repeat this step until all of the $k$ initial centroids $\tilde{\bm c}_1,\dots,\tilde{\bm c}_k$ are chosen.}  
\State{Let ${\bm c}_r=\tilde {\bm c}_r$ for all $r=1,\dots,k$.}
\Statex{\it Begin Iteration}
\State{For each ${\bm x}_i\in{\cal D}$, assign $r$ as the label of ${\bm x}_i$ if $\|{\bm x}_i-{\bm c}_r\|<\|{\bm x}_i-{\bm c}_{r'}\|$ for any $r'\not=r$.}
\State{For each $r=1,\dots,k$, update ${\bm c}_r$ by the average of data points with labels $r$.}
\Statex{\it End Iteration}
\State {Output}
\end{algorithmic}
\end{algorithm}

The final solution given by Algorithm~\ref{alg:improved k-means algorithm} depends on the first initial centroid as it is obtained by randomness. The main difference between Algorithm~\ref{alg:improved k-means algorithm} and other $k$-means algorithms is the initialization stage given by Steps 1 to 3. The goal is to make the minimum distance between the initial centroids as large as possible. The iteration stage given by Steps 5 and 6 is identical to that used by other $k$-means methods. Similarly, we cannot guarantee that the solution provided by Algorithm~\ref{alg:improved k-means algorithm} is a global solution of~\eqref{eq:k-means criterion}. We can only show that it is a local solution. We show that any local optimizer provided by Algorithm~\ref{alg:improved k-means algorithm} is clustering consistent, implying that local optimization is good enough if Algorithm~\ref{alg:improved k-means algorithm} is used.

\begin{thm}
\label{thm:local optimizer}
Any solution provided by Algorithm~\ref{alg:improved k-means algorithm} is a local optimizer of~\eqref{eq:k-means criterion}.
\end{thm}

\begin{lem}
\label{lem:initial centroids different clusters}
If Conditions (a), (c), and (d) of Theorem~\ref{thm:clustering consistency of the k means given k} are satisfied, then $\tilde{\bm c}_1,\dots,\tilde{\bm c}_k$ reside in distinct clusters in probability.
\end{lem}

\begin{thm}
\label{thm:local optimizer clustering consistency}
If Conditions (a), (c), and (d) of Theorem~\ref{thm:clustering consistency of the k means given k} are satisfied, then any local optimizer provided by Algorithm~\ref{alg:improved k-means algorithm} for~\eqref{eq:k-means criterion} is clustering consistent.
\end{thm}

\begin{cor}
\label{cor:local clustering consistency}
Assume that Condition (a) of Theorem~\ref{thm:clustering consistency of the k means given k} is satisfied, and $p$ is a constant. If  Condition (e) of Corollary~\ref{cor:clustering consistent sphere constant sigmasq and p} holds, then any local optimizer provided by Algorithm~\ref{alg:improved k-means algorithm} for~\eqref{eq:k-means criterion} is clustering consistent.
\end{cor}

Because any local optimizer given by Algorithm~\ref{alg:improved k-means algorithm} is clustering consistent, it does not matter whether the global optimizer of~\eqref{eq:k-means criterion} can be numerically computed in polynomial times or not. It is enough to use a local optimizer to partition the data set. This means that our research successfully reduces the $k$-means clustering problem to a local optimization problem, imply that the NP-hardness issue does not affect the implementation of our method. Therefore, our method is computationally efficient.

Corollary~\ref{cor:local clustering consistency} only requires that the minimum distance between the centroids of the clusters  approaches infinite with the speed faster than $\log^{1/2}{n}$. It does not matter whether $k$ is bounded or approaches infinite with $n$. Therefore, Algorithm~\ref{alg:improved k-means algorithm} can be implemented even when there exist many small clusters. Because outliers can be treated as clusters with sizes equal to $1$, our method can also be used even when outliers are present. This cannot be achieved by other $k$-means methods. We put this issue in Section~\ref{sec:simulation}.

Although we assume that $k$ is known in Algorithm~\ref{alg:improved k-means algorithm}, our method can also be used even when $k$ is unknown. The idea is to implement  Algorithm~\ref{alg:improved k-means algorithm} to various candidate values of $k$ with the best value of $k$ to be chosen by a predefined criterion. A few predefined criteria for $k$ have been proposed in the literature. Examples include the minimum message length (MML) criterion~\cite{figueiredo2002}, the minimum description length (MDL) criterion~\cite{hansen2001}, and the Bayesian information criterion (BIC)~\cite{zhanglin2021}. Besides, the Gap statistics is considered as one of the most popular approaches for the number of clusters~\cite{tibshirani2001}. Thus, we study the Gap statistics approach. We find that it works well  in our method but not in other $k$-means methods. We also put this issue in Section~\ref{sec:simulation}.

\section{Extension}
\label{sec:extension}

We extend our method to the  GMM defined by~\eqref{eq:underlying distribution of GMM} with arbitrary ${\bm\Sigma}_r$ and $\pi_r$ satisfying $\sum_{r=1}^k \pi_r=1$. Besides the $k$-means model, this also includes the linear discriminant analysis (LDA) model by assuming that all ${\bm\Sigma}_r$ are identical and the quadratic discriminant analysis (QDA) model by assuming that ${\bm\Sigma}_r$ for $r=1,\dots,k$ are different. The GMM has many other variants. Examples include the spherical and unequal volume model by setting ${\bm\Sigma}_r=\sigma_r^2{\bf I}$ with distinct $\sigma_r^2$, the diagonal and equal volume and shape model by setting ${\bm\Sigma}_r$ all identical and diagonal, and elliptical and equal volume and shape model by setting ${\bm\Sigma}_r=\sigma_r^2{\bm\Sigma}_0$ for a certain ${\bm\Sigma}_0$ with distinct $\sigma_r^2$. These variants have been incorporated in the \textsf{mclust} package of \textsf{R}. 

We show that our method is robust to the specifications of ${\bm\Sigma}_r$. We do not expect that any $k$-means method is criterion consistent due to violations of the assumptions of the $k$-means model. However, we can still show that our method is clustering consistent under a few weak regularity conditions. This means that criterion consistency and clustering consistency are two different concepts. Thus, they should be evaluated separately. 

We still focus on the clustering consistency problem of the proposed $k$-means method. To control the variations of variance-covariance matrices among clusters, we control the largest eigenvalue of ${\bm\Sigma}_r$. We show that the role of the largest eigenvalue used in this section is equivalent to the role of $\sigma^2$ used in the previous section, leading to clustering consistency of our $k$-means method for the general GMMs. 

\begin{thm}
\label{thm:clustering consistency of kmeans QDA}
Assume that ${\bm x}_1,\cdots,{\bm x}_n$ are independently collected from~\eqref{eq:underlying distribution of GMM} and each cluster has its own ${\bm\Sigma}_r$. If $\mathop{\lim\!\inf}_{n\rightarrow\infty} d/(32\lambda_{\max}\log{n})^{1/2}>1$ and $\lim_{n\rightarrow\infty} p\log[d^2/(p\lambda_{\max})]/\log{n}=0$, where $\lambda_{\max}$ is the maximum of the largest eigenvalue of ${\bm\Sigma}_r$ for $r=1,\dots,k$, then any local solution provided by Algorithm~\ref{alg:improved k-means algorithm} for~\eqref{eq:k-means criterion} is clustering consistent.
\end{thm}

\begin{cor}
\label{cor:clustering consistency of kmeans QDA, p fixed}
Suppose that the assumptions of Theorem~\ref{thm:clustering consistency of kmeans QDA} are satisfied and $p$ is fixed as $n\rightarrow\infty$. If (i) $\lambda_{\max}$ is fixed and $\mathop{\lim\!\inf}_{n\rightarrow\infty} d/(32\log{n})^{1/2}>\lambda_{\max}$ or (ii) $d$ is fixed and $\mathop{\lim\!\sup}_{n\rightarrow\infty}\lambda_{max}\log{n}<d^2/32$, then then any local solution provided by Algorithm~\ref{alg:improved k-means algorithm} for~\eqref{eq:k-means criterion} is clustering consistent.
\end{cor}

We have shown that our proposed $k$-mean method is clustering consistent even when ${\bm\Sigma}_r$ for $r=1,\dots,k$ are distinct. Thus, our method can be used to the general GMM defined by~\eqref{eq:underlying distribution of GMM} for data clustering. This includes the LDA and QDA models. As we do not specify any conditions for $k$ and $\pi_r$, our method can be implemented even if the cluster sizes are very unbalanced. This includes the case when some of the clusters are composed by one observation, meaning that they are outliers. If $k$ is unknown, we also use the Gap statistic to determine the number of clusters. We evaluate these issues in our simulation studies.

\section{Simulation}
\label{sec:simulation}

We compared our $k$-means method with the comparison to our competitors via Monte Carlo simulations. Our competitors included the EM-algorithm for GMMs clustering (denoted by\textsf{ EM-GMM}), the basic $k$-means clustering (using randomly selected initial cluster centroids), and the $k$-means++ clustering~\cite{arthur2007}. To make initial centroids far away from each other, the $k$-means++ selects the first initial centroid randomly from the data points. From the second initial centroid, it defines a distance measure as the minimum distance to the previous chosen centroids. The probability for the remaining data points to be selected as a new initial centroid is inversely proportional to the square of the distance measure. The procedure is repeated until all of the $k$ initial centroids are chosen. The remaining steps of the $k$-means++ are identical to those of the basic $k$-means.

Both the basic $k$-means and the $k$-means++ have four variants. The four variants of the basic $k$-means are the Hartigan-Wong $k$-means~\cite{hartigan1979}, the Lloyd $k$-means~\cite{lloyd1982}, the Forgy $k$-means~\cite{forgy1965}, and the MacQueen $k$-means~\cite{macqueen1967}. They are denoted as \textsf{HW}, \textsf{Lloyd}, \textsf{Forgy}, and \textsf{MacQueen}, respectively.  The four variants of the $k$-means++ are derived by the combinations of those with the $k$-means++ initialization, respectively. They are denoted as \textsf{HW++}, \textsf{Lloyd++}, \textsf{Forgy++}, and \textsf{MacQueen++}, respectively. We implemented the \textsf{EM-GMM} by the \textsf{Mclust} function of the \textsf{mclust} package, the four variants of the basic $k$-means by the \textsf{kmeans} function of the \textsf{base} library, the four variants of the $k$-means++ by the \textsf{kmeanspp} function of the \textsf{LICORS} package of \textsf{R}. Therefore, we had $10$ methods  in our simulation. The first was our proposed method. The remaining nine was our competitors. We compared those methods in the cases when $k$ was either known or unknown. When $k$ was unknown, we used the Gap statistics~\cite{tibshirani2001} to estimate the number of clusters. 

\begin{table}
\footnotesize
\caption{\label{tab:clustering error with and without outlier}  Simulations with $1000$ replications for the percentages of the clustering errors (i.e., $100{\rm CER}$) reported by the proposed $k$-means method with the comparison to its competitors in the cases when outliers are absent ($k=10$) and present ($k=20$ and $k_{outlier}=10$), respectively.}
\begin{center}
\begin{tabular}{ccccccc}\hline
 & \multicolumn{6}{c}{$k$-means Model}\\\cline{2-7}
& \multicolumn{3}{c}{$\phi$ for Outliers Absent} & \multicolumn{3}{c}{$\phi$ for Outliers Present}  \\\cline{2-7}
Method &  0.4 & 0.6 & 0.8 & 0.4 & 0.6 & 0.8 \\\hline
Proposed &$6.2$&$1.1$&$0.4$&$6.1$&$1.4$&$0.6$\\
\textsf{EM-GMM} &$12.0$&$10.7$&$6.1$&$36.8$&$33.7$&$21.6$\\
\textsf{HW} &$20.9$&$20.9$&$21.7$&$50.9$&$49.9$&$50.3$\\
\textsf{Lloyd} &$23.1$&$22.0$&$22.2$&$51.8$&$50.7$&$50.4$\\
\textsf{Forgy} &$22.7$&$22.5$&$21.8$&$51.9$&$50.6$&$50.3$\\
\textsf{MacQueen} &$21.3$&$21.5$&$22.2$&$51.3$&$50.4$&$50.8$\\
\textsf{HW++} &$16.0$&$15.4$&$15.3$&$44.8$&$45.1$&$45.4$\\
\textsf{Lloyd++} &$16.5$&$15.4$&$15.8$&$45.0$&$44.9$&$46.0$\\
\textsf{Lloyd++} &$16.6$&$16.0$&$15.8$&$45.6$&$45.3$&$45.8$\\
\textsf{MacQueen++} &$16.3$&$16.2$&$15.7$&$45.3$&$45.0$&$46.0$\\\hline\hline
 & \multicolumn{6}{c}{QDA Model }\\\cline{2-7}
Method & \multicolumn{3}{c}{ $\phi$ for Outliers Absent} & \multicolumn{3}{c}{$\phi$ for Outliers Present}  \\\hline
Proposed  &$10.4$&$3.3$&$0.8$&$10.4$&$3.0$&$1.0$\\
\textsf{EM-GMM} &$20.1$&$16.6$&$11.0$&$48.4$&$45.4$&$41.8$\\
\textsf{HW} &$23.1$&$21.4$&$21.6$&$50.4$&$48.5$&$48.5$\\
\textsf{Lloyd} &$25.2$&$22.8$&$22.2$&$51.6$&$49.5$&$49.3$\\
\textsf{Forgy} &$25.2$&$22.9$&$22.3$&$51.0$&$49.4$&$49.2$\\
\textsf{MacQueen} &$23.4$&$22.4$&$21.9$&$50.7$&$49.0$&$49.0$\\
\textsf{HW++} & $19.0$&$16.0$&$15.7$&$43.8$&$42.4$&$42.6$\\
\textsf{Lloyd++} &$19.8$&$16.3$&$15.9$&$43.1$&$42.9$&$43.2$\\
\textsf{Forgy++} &$19.7$&$16.2$&$15.7$&$43.3$&$42.7$&$43.4$\\
\textsf{MacQueen++} &$19.8$&$16.7$&$16.0$&$43.4$&$43.0$&$43.2$\\\hline
\end{tabular}
\end{center}
\end{table}
\normalsize

We used the GMM defined by~\eqref{eq:underlying distribution of GMM} with $p=5$ to generate clusters. We considered two scenarios. In the first scenario, we chose ${\bm\Sigma}_r=\sigma^2{\bf I}$ for all $r=1,\dots,k$, implying that it was the $k$-means model. In the second scenario, we chose all ${\bm\Sigma}_r$ different, implying that it was the QDA model. In both scenarios, we compared the ten methods under the cases when outliers were absent and present, respectively. 

We carried out a three-step procedure to generate the clusters. In the first step, we generated the cluster expected vectors ${\bm\mu}_r$ independently by ${\cal N}({\bm 0},\phi^2{\bf I})$. In the second step, we generated the clusters sizes $n_r$. When the outliers were absent, we chose $k=10$. When the outliers were present, we chose $k=20$. In both cases, we generated $5$ small clusters and $5$ large clusters. If outliers were present, then the number of outliers was $k_{outlier}=10$.  The sizes of the small clusters were independently generated by $n_r~\sim{\cal P}(50)$. The sizes of the large clusters were independently generated by $n_r\sim{\cal P}(1000)$. The cluster sizes for outliers were $n_r=1$. In the third step, we  generated the observations. In the $k$-means model, we independently generated $n_r$ observations from ${\cal N}({\bm\mu}_r,0.1^2{\bf I})$. In the QDA model, for the $r$th cluster, we first independently generated $5$ eigenvalues $d_{r1},\dots,d_{r5}$ of ${\bm\Sigma}_r$ by $d_{rj}\sim{\cal U}(0,0.2)$ for $j=1,\dots,5$. We then independently generated $5$ eigenvectors of ${\bm\Sigma}_r$ by a random orthogonal matrix ${\bf U}_r$. We calculated the variance-covariance matrix of the $r$th cluster by ${\bm\Sigma}_r={\bf U}_r{\bf D}_r{\bf U}_r^\top$ with ${\bf D}_r={\rm diag}(d_{r1},\dots,d_{r_5})$. After all ${\bm\Sigma}_r$ were derived, we independently generated $n_r$ observations of the $r$th cluster by ${\cal N}({\bm\mu}_r,{\bm\Sigma}_r)$. The total number of observations was $n=\sum_{r=1}^k n_r$. 

We used the percentages of cluster errors to compare our proposed method with our competitors. Due to the permutation issue, the clustering labels are equivalent if the result of one method can be changed to that of the other method by a permutation $\tau$ on $\{1,\dots,k\}$. Using this criterion, we compute the percentage of clustering errors. In particular, let $\{\hat n_1,\dots,\hat n_k\}$ be the sizes of clusters given by a clustering method. The estimated labels may not match the true labels. To address the issue, we fix the true labels and change the estimated labels by a permutation, leading to the clustering error rate as
\begin{equation}
\label{eq:clustering error couut}
{\rm CER}=1-{1\over n}\max_{\tau}\sum_{r=1}^k \hat n_{r,\tau(r)},
\end{equation}
where $\hat n_{r,\tau(r)}$ is the number of data points with label $\tau(r)$ assigned by the clustering method and label $r$ assigned by the truth.  The percentage of the clustering errors is $100{\rm CER}$. The computation for ${\rm CER}$ can be easily solved by the well known Hungarian algorithm. 

We generated $1,\!000$ data sets from~\eqref{eq:underlying distribution of GMM} for each selected $\phi$ in the $k$-means and the QDA models, respectively. We applied all of the ten methods to each data set. We computed the percentages of clustering errors (Table~\ref{tab:clustering error with and without outlier}). We compared the performance of the ten methods under the $k$-means and the QDA models without and with the outliers, respectively. We found that our method was consistently and significantly better than our competitors. In the comparison between our competitors, we found that the \textsf{EM-GMM} method was better than the four basic $k$-means and the four $k$-means++ methods. There were no significantly differences within the four basic $k$-means and the four $k$-means++ methods, respectively.  The four $k$-means++ methods were better than the four basic $k$-means methods. This supports the previous statement that the initialization adopted by the $k$-means++ improves $k$-means clustering. Because the $k$-means++ was significantly worse than our proposed method, we conclude that the initialization adopted by the $k$-means++ is far from the optimal. The proposed method significantly improves the quality of $k$-means clustering.

We then compared the performance of each of ten methods between the $k$-means and the QDA models in the cases when outliers were absent and present, respectively. We found that the performance of our method was not significantly affected by the presence of the outliers, but the performance of our competitor was. This means that our method is robust to outliers, but our competitors are not. In the QDA model, the performance of our method was still better than that of the \textsf{EM-GMM} method, implying that the proposed $k$-means method is also appropriate for general GMMs. 

\begin{table}
\footnotesize
\caption{\label{tab:outlier deteection}  Simulations with $1000$ replications for the average number of outliers reported by the proposed $k$-means method with the comparison to its competitors, where $k=20$ and $k_{outlier}=10$.}
\begin{center}
\begin{tabular}{ccccccc}\hline
 & \multicolumn{3}{c}{$\phi$ for $k$-means Model} & \multicolumn{3}{c}{$\phi$ for QDA Model} \\\cline{2-7}
Method & $0.4$ & $0.6$ & $0.8$ & $0.4$ & $0.6$ & $0.8$\\\hline
Proposed  &$9.94$&$9.93$&$9.94$&$9.97$&$9.97$&$9.98$\\
\textsf{EM-GMM}&$0.00$&$0.00$&$0.00$&$0.00$&$0.00$&$0.00$\\
\textsf{HW}&$0.07$&$0.06$&$0.10$&$0.20$&$0.25$&$0.27$\\
\textsf{Lloyd}&$0.17$&$0.16$&$0.16$&$0.31$&$0.37$&$0.35$\\
\textsf{Forgy}&$0.14$&$0.17$&$0.16$&$0.31$&$0.35$&$0.34$\\
\textsf{MacQueen}&$0.21$&$0.19$&$0.18$&$0.45$&$0.44$&$0.44$\\
\textsf{HW++}&$0.38$&$0.23$&$0.18$&$0.83$&$0.63$&$0.53$\\
\textsf{Lloyd++}&$0.81$&$0.53$&$0.36$&$1.67$&$1.15$&$0.89$\\
\textsf{Forgy++}&$0.84$&$0.50$&$0.32$&$1.63$&$1.16$&$0.85$\\
\textsf{MacQueen++}&$0.85$&$0.53$&$0.37$&$1.68$&$1.17$&$0.88$\\\hline
\end{tabular}
\end{center}
\end{table}
\normalsize

We studied whether the ten methods could be used to identify outliers when they were present. We claimed a cluster as an outlier if its size was one.  The simulation results  showed that only our method successfully identified all of the outliers. All of our competitors failed to identify most of the outliers. In all of the cases that we simulated, the \textsf{EM-GMM} did not discover any outliers. In most of the cases, the basic $k$-means the $k$-means++ methods identified zero or one outlier. Therefore, we conclude that the proposed method can identify outliers but our competitors cannot.

\begin{figure}
\centerline{\rotatebox{270}{\psfig{figure=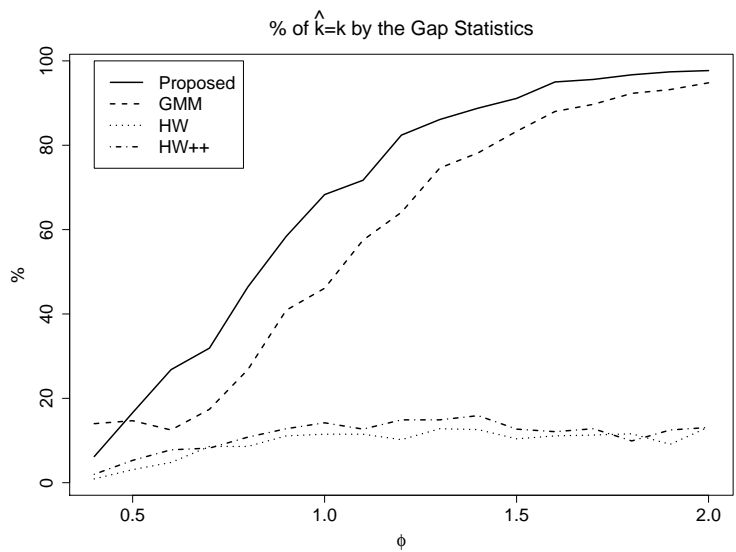,width=3.0in,}}}
\caption{\label{fig:estimation of k gap}Percentage of correctly identified number of clusters (i.e., \% of $\hat k=k$) reported by the combination of the Gap statistics with the proposed $k$-means, the \textsf{EM-GMM}, the \textsf{HW}, and the \textsf{HW++} methods, respectively.}
\end{figure}

In the end, we compared the performance of our method with that of our competitors in the case when $k$ was unknown. In practice, the determination of the number of clusters is considered as one of the most important issues for the $k$-means. Many approaches have been proposed in the literature. Examples include the MML criteria~\cite{figueiredo2002}, the MDL criterion~\cite{hansen2001}, the BIC~\cite{zhanglin2021}, and the Bayesian non-parametric~\cite{ferguson1973,rasmussen1999} approaches. Among those, the Gap statistics~\cite{tibshirani2001} is considered as one of the most popular approaches. Therefore, we decided to carry out the comparison based on the Gap statistics. In particular, we estimated the number of clusters by the combination of each of the ten methods with the Gap statistic. We denoted $\hat k$ as the estimated number of clusters and $k$ as the true number of clusters. We concluded that the true number of clusters was correctly identified if $\hat k=k$ or misidentified otherwise. We compared the ten methods using the percentage of $\hat k=k$ (Figure~\ref{fig:estimation of k gap}).  Because there was no significant difference among the four basic $k$-means and the four $k$-means++ methods, we only report the results for \textsf{HW} and \textsf{HW++}. Our simulations showed that both the proposed $k$-means and the \textsf{EM-GMM} colud identify the true number of clusters if they were combined with the Gap statistics. None of the basic $k$-means or $k$-means++ methods could identify the number of clusters. 

In summary, our simulations show that our proposed $k$-means method outperforms our competitors in all of the aspects that we have studied. It has low clustering error rates. It can also detect outliers. Our method can identify the number of clusters in the case when the true $k$ is unknown. Our study generally supports the Gap statistics approach for the determination of the number of clusters.  It strongly suggests not to implement the Gap statistics approach to the basic $k$-means or $k$-means++ methods for the determination of the number of clusters.

\section{Application}
\label{sec:application}

We implemented our method with the comparison to our competitors to the Swiss Banknote data set. The data set was originally provided by Tables 1 and 2 on Page 5--8 of~\cite{flury1988}. It is now available in the \textsf{mclust} package of~\textsf{R}. The Swiss Banknote data set contains six measurements (denoted by $x_1,\dots,x_6$, respectively) made on $100$ genuine and $100$ counterfeit old-Swiss 1000-franc bank notes of the Swiss National Bank in 1980s. Thus, $n=200$. It also contains an additional \textsf{status} variable for the status of the banknotes as \textsf{genuine} or \textsf{counterfeit}. Thus, \textsf{status} is the ground truth variable. It was used in the computation of CER.

\begin{figure}
\centerline{\rotatebox{270}{\psfig{figure=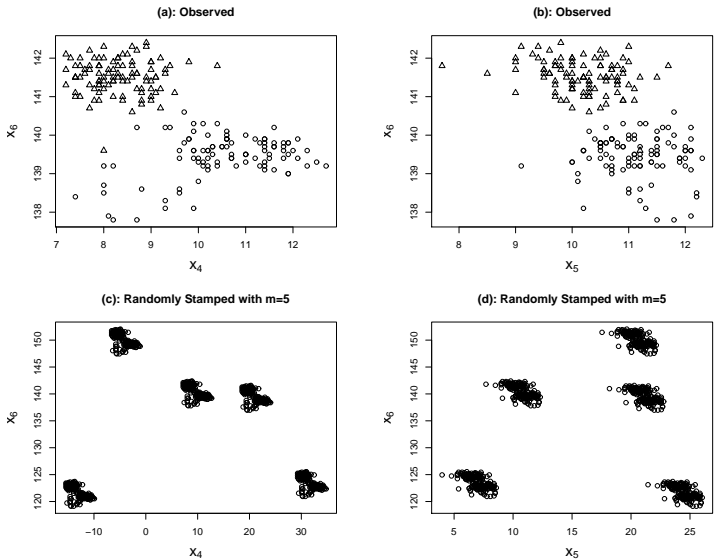,width=3.0in,}}}
\caption{\label{fig:banknote generated clusters} The original observed cluster pattern by plots (a) $x_4$ vs $x_6$ and (b) $x_5$ vs $x_6$, and the generated randomly stamped cluster pattern displayed by plots (c) $x_4$ vs $x_6$ and (d) $x_5$ vs $x_6$.}
\end{figure}

We studied two circumstances. In the first circumstance,  we implemented all of the ten methods directly to the observed data. We computed the percentage of cluster errors (i.e., $100{\rm CER}$) for each method using~\eqref{eq:clustering error couut}, where we assumed that $k=2$ was known. The percentage of the proposed method, the basic \textsf{HW}, \textsf{Loyod}, \textsf{Forgy}, and \textsf{MacQueen} methods, and  the \textsf{HW++}, \textsf{Loyod++}, \textsf{Forgy++}, and \textsf{MacQueen++} methods  were all $0\%$.  The percentage of the {\textsf{EM-GMM} was $0.5\%$. Therefore, we conclude that the performances of all of the methods were close in the direct implementation to the Swiss Banknote data set.

To emphasize the importance of clustering consistency, we studied the second circumstance. We used the random stamps approach. The random stamps approach treats the observed pattern as the base stamp. The base stamp is duplicated first and them shifted randomly by a distribution. In the end, the random stamps approach generates a pattern by duplicating the base stamp with a number of location shifts. The goal for us to consider the random stamps approach was to investigate the influence of well separated cluster patterns on a selected clustering method. 

To implement the random stamps approach, we first selected positive integer $m$ for the number of stamps. We then generated $m$ vectors ${\bm\xi}_v=(\xi_{v1},\dots,\xi_{v6})$ identically and independently from the multivariate normal distribution with mean zero and variance-covariance matrix ${\bf A}={\rm diag}(10s_1,\dots,10s_6)$, where $s_v$  was the standard error of $x_j$ for $j=1,\dots,6$ provided by the base stamp. We used ${\bm\xi}_v$ to shift the base stamp pattern individually. The goal was to ensure that all of the stamps were well separated by the location shifts. To make it clearer, we illustrate a realization of the random stamps approach by Figure~\ref{fig:banknote generated clusters} with $m=5$ stamps. Note that the Swiss Banknote data set contains $6$ variables. We need to generate a six-dimensional vector to shift the base stamp each time. Based on the original observed cluster patterns displayed by the plot using $x_4$ and $x_6$ given by Figure~\ref{fig:banknote generated clusters}(a), we shifted $x_{i4}$ to $x_{i4}+\xi_{v4}$ and $x_{i6}$ to $x_{i6}+\xi_{v6}$, $i=1,\dots,n$. We obtained Figure~\ref{fig:banknote generated clusters}(c). Similarly, we obtained Figure~\ref{fig:banknote generated clusters}(d) using Figure~\ref{fig:banknote generated clusters}(b). Figures~\ref{fig:banknote generated clusters}(c) and~\ref{fig:banknote generated clusters}(d) were obtained from the same data set using different variables for visualization. Thus, the number of clusters was $k=2m=10$. The stamps were well separated. Each pair of clusters contained by a stamp was identical to the observed cluster pattern contained by the base stamp. Therefore, the performance of a nice clustering method should not be significantly  influenced by the random stamps approach used in the application. 

\begin{figure}
\centerline{\rotatebox{270}{\psfig{figure=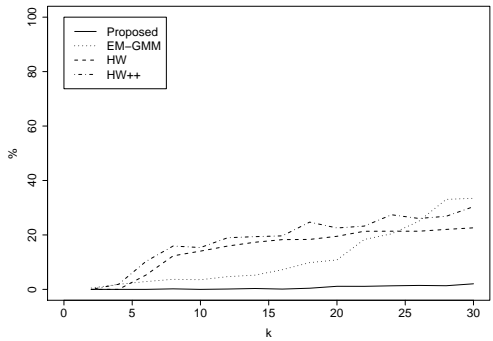,width=2.0in,}}}
\caption{\label{fig:banknote percentage of cluster error duplicated} Percentage (\%) of cluster errors for the generated patterns derived by randomly stamping the Swiss Banknote data set, where the number of clusters is $k=2m$.}
\end{figure}

We varied $m$ from $1$ to $15$ and obtained $k=2$ to $k=30$  clusters, respectively, by the random stamps approach for the Swiss Banknote data. We implemented the proposed $k$-means, the four basic $k$-means, the four $k$-means++, and the \textsf{EM-GMM} methods. We compared these methods using the percentage of clustering errors (Figure~\ref{fig:banknote percentage of cluster error duplicated}).  As the shifts were randomly generated, we repeated $1000$ times for each selected $m$. As the curves for the four basic $k$-means and the four $k$-means methods were close, respectively, we only put the curves for the \textsf{HW} basic $k$-means and \textsf{HW++} $k$-means++ methods in Figure~\ref{fig:banknote percentage of cluster error duplicated}. We found that only the percentages of clustering errors reported by the proposed $k$-means method were consistently low. This means that our competitors are not reliable to the increase of the number of clusters even if they are far away from each other. This problem has been overcome by our method. Therefore, we conclude that the performance of the proposed $k$-means method is better than our competitors in the application to the Swiss Banknote data set.

\section{Discussion}
\label{sec:discussion}

We study asymptotic properties for the $k$-means clustering. We point out that the proposed clustering consistency is more appropriate than the previous criterion consistency for the $k$-means, when it is applied to the recent unsupervised learning problems arisen in statistics, machine learning, and artificial intelligence. Using the proposed clustering consistency, we investigate existing $k$-means methods, including the basic $k$-means and the $k$-means++ adopted by many software packages, such as \textsf{R} and \textsf{python}. We find that all of the existing $k$-means methods are not optimal. To ensure a $k$-means method to be clustering consistent, it is more appropriate to choose the initial $k$-centroids as far as possible, leading to the proposed $k$-means method. We show that any local solution provided by the proposed $k$-means method is clustering consistent, implying that it is better than the existing $k$-means methods adopted by \textsf{R} and \textsf{python}. More interesting, we find that the combination of the $k$-means with the well known Gap statistics can be used to determine the number of clusters when $k$ is unknown, which cannot be achieved by the basic $k$-means and the $k$-means++ methods.

The performance of the basic $k$-means and the $k$-means++ becomes worse as $k$ becomes large. This can be interpreted by Theorem~\ref{thm:clustering consistency of the k means given k} of the article. For any $k$-means method, if it can guarantee that the centroids reside in distinct clusters sooner or later in the iterations, then the $k$-means method is clustering consistent. If a basic $k$-means or a $k$-means++ method is used to the case when $k$ is small (e.g., $k=2$ or $3$), then the centroids may belong to distinct clusters after a few steps of iterations. This can be hardly achieved in the case when $k$ is large (e.g., $k=10$ or higher). Therefore, it is more important to evaluate a clustering method under the case when $k$ is not very small.  In addition, we find that the performance of the proposed $k$-means is better than that of the \textsf{EM-GMM} in all of the cases that we have  studied, implying that the proposed $k$-means is comparable to the GMM method. If a basic $k$-means or a $k$-means++ method is used, then we cannot draw a similar conclusion. Therefore, the derivation of an optimal $k$-means method is critical in the comparison between the $k$-means and its competitors.

We treat clustering consistency as the minimum requirement of a clustering method. It means that if the distances between individual clusters increases, then the clustering error rates should be  small.  We show that clustering consistency can be achieved by local solutions of our method, indicating that the NP-hardness is not an issue in our method. In practice, it is enough to use our method to provide a local optimizer. Note that clustering consistency is an asymptotic property and the sample size is always finite in practice. We may still gain benefits if we want to determine a better solution by replications. However, this does not affect the asymptotic properties formulated under clustering consistency proposed in the current research.



\appendix
\section{Proofs}

{\bf Proof of Theorem~\ref{thm:clustering consistency of the k means given k}.} Let $\tilde{\bm c}_1,\dots,\tilde{\bm c}_k$ be the initial centroids of the clusters selected by the $k$-means. Without the loss of generality, we can assume that their true cluster labels are $1,\cdots,k$, respectively. In the first iteration, the $k$-means assigns ${\bm x}_i$ to the $r$th cluster if $\|{\bm x}_i-\tilde{\bm c}_{r}\|<\|{\bm x}_i-\tilde{\bm c}_{r'}\|$ for any $r'\not=r$. Let $\check{r}_i$  be the true cluster label of ${\bm x}_i$. By the implmentation of the triangle inequality to  $\|{\bm x}_i-\tilde{\bm c}_{r}\|=\|({\bm x}_i-{\bm\mu}_{\check{r}_i})+({\bm\mu}_{\check{r}_i}-{\bm\mu}_r)+({\bm\mu}_r-\tilde{\bm c}_{r})\|$, we have 
$$\eqalign{
\|{\bm x}_i-\tilde{\bm c}_{r}\|\ge & \|{\bm\mu}_{\check{r}_i}-{\bm\mu}_r\|-\|({\bm x}_i-{\bm\mu}_{\check{r}_i})+({\bm\mu}_r-\tilde{\bm c}_{r})\|\cr
\ge & \|{\bm\mu}_{\check{r}_i}-{\bm\mu}_r\|-(\|{\bm x}_i-{\bm\mu}_{\check{r}_i}\|+\|\tilde{\bm c}_{r}-{\bm\mu}_r\|).\cr
}$$ 
Similarly, by $\|{\bm x}_i-\tilde{\bm c}_{\check {r}_i}\| =\|({\bm x}_i-{\bm\mu}_{\check{r}_i})+({\bm\mu}_{\check{r}_i}-\tilde{\bm c}_{\check {r}_i})\|$, we have
$$
\|{\bm x}_i-\tilde{\bm c}_{\check {r}_i}\| \le \|{\bm x}_i-{\bm\mu}_{\check{r}_i}\|+\|\tilde{\bm c}_{\check {r}_i}-{\bm\mu}_{\check{r}_i}\|.
$$
If $r\not=\check{r}_i$, then $\|{\bm x}_i-\tilde{\bm c}_{r}\|<\|{\bm x}_i-\tilde{\bm c}_{\check {r}_i}\| $, leading to 
$$\eqalign{
\|{\bm x}_i-{\bm\mu}_{\check{r}_i}\|\ge & {1\over 2}\|{\bm\mu}_{\check{r}_i}-{\bm\mu}_r\| -{1\over 2}( \|\tilde{\bm c}_{\check {r}_i}-{\bm\mu}_{\check{r}_i}\|+\|\tilde{\bm c}_{r}-{\bm\mu}_r\|)\cr
\ge & {d\over 2}-{1\over 2}( \|\tilde{\bm c}_{\check {r}_i}-{\bm\mu}_{\check{r}_i}\|+\|\tilde{\bm c}_{r}-{\bm\mu}_r\|).\cr
}$$
It is derived by the combination of the above two inequalities with the definition of $d$ given by~\eqref{eq:minium distance}. Note that all of $\|{\bm x}_i-{\bm\mu}_{\check{r}_i}\|^2$,  $\|\tilde{\bm c}_{\check{r}_i}-{\bm\mu}_{\check{r}_i}\|^2$, and $\|\tilde{\bm c}_{r}-{\bm\mu}_r\|^2$ are $\sigma^2\chi_{p}^2$ random variables. We can use the formulation for the upper tail probability of the $\chi_q^2$-distribution given by~\cite{inglot2010} as
\begin{equation}
\label{eq:upper tail chi-square distribution}
{\rm Pr}\{\chi_q^2\ge u\}\le {u\over\sqrt{\pi}(u-q+2)}\exp\left\{-{1\over 2}[u-q-(q-2)\log(u/q)+\log{q}]\right\},
\end{equation}
for any $q\ge 2$ and $u\ge q-2$. If $p\ge 2$, then 
$$\eqalign{
&{\rm Pr}({\bm x}_i\not\in C_{r,T}\ \exists i\in\{1,\dots,n\})\cr
\le &\sum_{i=1}^n {\rm Pr}({\bm x}_i\not\in C_{r,T})\cr
=&\sum_{i=1}^n {\rm Pr}\left\{ \|{\bm x}_i-{\bm\mu}_{\check{r}_i}\| + {1\over 2} \|\tilde{\bm c}_{\check {r}_i}-{\bm\mu}_{\check{r}_i}\|+{1\over 2}\|\tilde{\bm c}_{r}-{\bm\mu}_r\| \ge {d\over 2} \right\}\cr
\le &\sum_{i=1}^n \left\{{\rm Pr}\left(\|{\bm x}_i-{\bm\mu}_{\check{r}_i}\|\ge {d\over 4}\right) + {\rm Pr}\left( \|\tilde{\bm c}_{\check {r}_i}-{\bm\mu}_{\check{r}_i}\|\ge {d\over 4}\right)+{\rm Pr}\left(\|\tilde{\bm c}_{r}-{\bm\mu}_r\| \ge {d\over 4}\right )\right\}\cr
\le & 3n {\rm Pr}(\chi_p^2\ge {d^2\over 16\sigma^2}).\cr
}
$$
By taking $q=p$ and $u=d^2/(16\sigma^2)$ in~\eqref{eq:upper tail chi-square distribution}, we have
\begin{equation}
\label{eq:upper probability kmeans spherical d>1}
\eqalign{
& {\rm Pr}({\bm x}_i\not\in C_{r,T}\ \exists i\in\{1,\dots,n\})\cr
\le & {3d^2\over\sqrt{\pi}[d^2-16\sigma^2(p-2)]}\exp\left\{-{1\over 2}\left[ {d^2\over 16\sigma^2}-p-(p-2)\log{d^2\over 16p\sigma^2}+\log{p}-2\log{n}  \right]\right\}.
}
\end{equation}
Using Condition $\lim_{n\rightarrow\infty} p\log[d^2/(p\sigma^2)]/\log{n}=0$,  we ignore the second, third, and fourth terms in the exponent when we evaluate the limit of the above probability. We only need to compare the first and the last terms. We find that the probability goes to $0$ if $\lim_{n\rightarrow\infty}[d^2/(16\sigma^2)-2\log{n}]=\infty$, which can be achieved by Condition $\lim\!\inf_{n\rightarrow\infty} d/[4\sigma(2\log{n})^{1/2}]>1$  of the theorem. If $p=1$, then by the same approach, we have
\begin{equation}
\label{eq:upper probability kmeans spherical d=1}
{\rm Pr}({\bm x}_i\not\in C_{r,T}\ \exists i\in\{1,\dots,n\})\le 3n{\rm Pr}(\chi_1^2\ge {d^2\over 16\sigma^2}).
\end{equation}
By 
$$\lim_{t\rightarrow\infty}{{\rm Pr}[{\cal N}(0,1)\ge t]\over ({\sqrt{2\pi}t})^{-1}e^{-{t^2\over 2}}}=1$$
and the connection between the $\chi_1^2$ and ${\cal N}(0,1)$ distributions, we have
$${\rm Pr}\left(\chi_1^2\ge {d^2\over 16\sigma^2}\right)\le e^{-{d^2\over 32\sigma^2}}$$
for sufficiently large $d^2/\sigma^2$. Then, we obtain 
$${\rm Pr}({\bm x}_i\not\in C_{r,T}\ \exists i\in\{1,\dots,n\})\le e^{-({d^2\over 32\sigma^2}-\log{n})}.$$
It also approaches $0$ as $n\rightarrow\infty$ under the assumptions of the theorem. Asymptotically, $\check{r}_i$ and $r$ must be identical  for all $i$. We draw the conclusion. \qed

{\bf Proof of Corollary~\ref{cor:clustering consistent sphere constant sigmasq and p}.} By treating $p$ as a constant in the evaluation of the asymptotic properties of~\eqref{eq:upper probability kmeans spherical d>1} and~\eqref{eq:upper probability kmeans spherical d=1} as $n\rightarrow\infty$, we can show that the $k$-means method is clustering consistent if $\lim\!\inf_{n\rightarrow\infty} d^2/(32\sigma^2\log{n})>1$, which is satisfied when either (i) or (ii) of (e) holds.  \qed

{\bf Proof of Theorem~\ref{thm:local optimizer}.} It is enough to show that the SSQ criterion always decreases in Steps 5 and 6, respectively, in Algorithm~\ref{alg:improved k-means algorithm}. In Step 5, each ${\bm x}_i$ is assigned a label by lowest $\|{\bm x}_i-{\bm c}_r\|$ value. The $SSQ(\widehat{\mathscr{C}}_{k})$ value does not increase. In Step 6, each ${\bm c}_r$ is the least square estimator of the cluster centroid based on the given label. The $SSQ(\widehat{\mathscr{C}}_{k})$ value does not increase either.  Therefore, any solution provided by Algorithm~\ref{alg:improved k-means algorithm} is a local optimizer of~\eqref{eq:k-means criterion}. \qed

{\bf Proof of Lemma~\ref{lem:initial centroids different clusters}.} Let $\tilde{\cal C}_{prev}=\{\tilde{\bm c}_1,\cdots,\tilde{\bm c}_{r-1}\}$ be the initial centroids selected by the previous steps and assume that the current step is to select $\tilde{\bm c}_r$. For distinct ${\bm x}_i,{\bm x}_{i'}\in{\cal D}\backslash\tilde{\cal C}_{prev}$ satisfying that $r$ does not reside in the clusters containing $\tilde{\cal C}_{prev}$ but $r'$ does, where the true labels of ${\bm x}_i$ and ${\bm x}_{i'}$ are $r$ and $r'$, respectively, we implement the triangle inequality to $\|{\bm x}_i-\tilde{\bm c}_{r'}\|=\|({\bm x}_i-{\bm\mu}_{r})+({\bm\mu}_{r}-{\bm\mu}_{r'})+({\bm\mu}_{r'}-\tilde{\bm c}_{r'})\|$ and obtain 
$$\|{\bm x}_i-\tilde{\bm c}_{r'}\|\ge \|{\bm\mu}_{r}-{\bm\mu}_{r'}\|-(\|{\bm x}_i-{\bm\mu}_{r}\|+\|{\bm\mu}_{r'}-\tilde{\bm c}_{r'}\|).$$
We implement the triangle inequality to $\|{\bm x}_{i'}-\tilde{\bm c}_{r'}\|=\|({\bm x}_{i'}-{\bm\mu}_{r'})+({\bm\mu}_{r'}-\tilde{\bm c}_{r'})\|$ and obtain
$$\|{\bm x}_{i'}-\tilde{\bm c}_{r'}\|\le \|{\bm x}_{i'}-{\bm\mu}_{r'}\|+\|{\bm\mu}_{r'}-\tilde{\bm c}_{r'}\|.$$
Combining the above two inequalities, we obtain 
$$\eqalign{
\|{\bm x}_i-\tilde{\bm c}_{r'}\|-\|{\bm x}_{i'}-\tilde{\bm c}_{r'}\|\ge & \|{\bm\mu}_{r}-{\bm\mu}_{r'}\|-(\|{\bm x}_i-{\bm\mu}_{r}\|+\|{\bm\mu}_{r'}-\tilde{\bm c}_{r'}\|\cr
&\hspace{2cm}+\|{\bm x}_{i'}-{\bm\mu}_{r'}\|+\|{\bm\mu}_{r'}-\tilde{\bm c}_{r'}\|),\cr
} $$
leading to 
$$\eqalign{
\|{\bm x}_i-\tilde{\bm c}_{r'}\|-\|{\bm x}_{i'}-\tilde{\bm c}_{r'}\|\ge & \min_{r\not=r'} \|{\bm\mu}_{r}-{\bm\mu}_{r'}\| -\max\{(\|{\bm x}_i-{\bm\mu}_{r}\|+\|{\bm\mu}_{r'}-\tilde{\bm c}_{r'}\|\cr
&\hspace{2cm}+\|{\bm x}_{i'}-{\bm\mu}_{r'}\|+\|{\bm\mu}_{r'}-\tilde{\bm c}_{r'}\|)\}\cr
=&d- \max\{(\|{\bm x}_i-{\bm\mu}_{r}\|+\|{\bm\mu}_{r'}-\tilde{\bm c}_{r'}\|\cr
&\hspace{1cm}+\|{\bm x}_{i'}-{\bm\mu}_{r'}\|+\|{\bm\mu}_{r'}-\tilde{\bm c}_{r'}\|)\}.\cr
}$$
Because $\|{\bm x}_i-{\bm\mu}_{r}\|^2$, $\|{\bm\mu}_{r'}-\tilde{\bm c}_{r'}\|^2$, $\|{\bm x}_{i'}-{\bm\mu}_{r'}\|^2$, and $\|{\bm\mu}_{r'}-\tilde{\bm c}_{r'}\|^2$ are $\sigma^2\chi_p^2$ random variables, we have
$$
\eqalign{
&{\rm Pr}\left[\min(\|{\bm x}_i-\tilde{\bm c}_{r'}\|-\|{\bm x}_{i'}-\tilde{\bm c}_{r'}\|)>0\right]\cr
\ge &1- {\rm Pr}\left[\max (\|{\bm x}_i-{\bm\mu}_{r}\|+\|{\bm\mu}_{r'}-\tilde{\bm c}_{r'}\|+\|{\bm x}_{i'}-{\bm\mu}_{r'}\|+\|{\bm\mu}_{r'}-\tilde{\bm c}_{r'}\|)>d  \right]\cr
\ge &1- \sum_{i=1}^n {\rm Pr}\left[ (\|{\bm x}_i-{\bm\mu}_{r}\|+\|{\bm\mu}_{r'}-\tilde{\bm c}_{r'}\|+\|{\bm x}_{i'}-{\bm\mu}_{r'}\|+\|{\bm\mu}_{r'}-\tilde{\bm c}_{r'}\|)> d  \right]\cr
\ge & 1-4n{\rm Pr}(\chi_p^2>{d^2\over 16\sigma^2}).\cr
}
$$
If $p\ge 2$, then by~\eqref{eq:upper tail chi-square distribution}, we obtain
\begin{equation}
\label{eq:minimum difference inequalities for the initial centroids by the algorithm}
\eqalign{
&{\rm Pr}\left[\min(\|{\bm x}_i-\tilde{\bm c}_{r'}\|-\|{\bm x}_{i'}-\tilde{\bm c}_{r'}\|)>0\right]\cr
\ge & 1-{4d^2\over\sqrt{\pi}[d^2-16\sigma^2(p-2)]}\exp\left\{ -{1\over 2}\left[{d^2\over 16\sigma^2}-p-(p-1)\log{d^2\over 16p\sigma^2}+\log{p}-2\log{n}\right] \right\},
}
\end{equation}
which goes to $1$ as $n\rightarrow\infty$. This means that ${\bm x}_{r'}$ cannot be the $r$ initial centroid asymptotically. Note that ${\bm x}_i$ and ${\bm x}_{i'}$ represent general points satisfying the condition. We draw the conclusion for the case when $p\ge 2$.  Similarly, we draw the conclusion for the case when $p=1$. This completes the proof.   \qed.

{\bf Proof of Theorem~\ref{thm:local optimizer clustering consistency}.} The conclusion can be directly implied by the Combination of Theorem~\ref{thm:clustering consistency of the k means given k} and Lemma~\ref{lem:initial centroids different clusters}. \qed

{\bf Proof of Corollary~\ref{cor:local clustering consistency}.} The conclusion can be directly implied by the Combination of Corollary~\ref{cor:clustering consistent sphere constant sigmasq and p} and Lemma~\ref{lem:initial centroids different clusters}. \qed

{\bf Proof of Theorem~\ref{thm:clustering consistency of kmeans QDA}.} We almost repeat the entire proofs of Theorems~\ref{thm:clustering consistency of the k means given k} and Lemma~\ref{lem:initial centroids different clusters} except the distribution of $\|{\bm x}_i-{\bm\mu}_{r_i}\|^2$ when the true label of ${\bm x}_i$ is $r_i$. As ${\bm\Sigma}_{r_i}$ is a general variance-covariance matrix of a multivariate normal distribution,  the distribution of $\|{\bm x}_i-{\bm\mu}_{r_i}\|^2$ cannot be expressed by a constant times a chi-square distribution. We only have $({\bm x}_i-{\bm\mu}_{r_i})^\top{\bm\Sigma}_{r_i}^{-1}({\bm x}_i-{\bm\mu}_{r_i})=\|{\bm z}_i\|^2\sim\chi_p^2$ for each $i\in\{1,\dots,n\}$, where ${\bm z}_i={\bm\Sigma}_r^{-1/2}({\bm x}_i-{\bm\mu}_{r_i})$. By ${\bm x}_i-{\bm\mu}_{r_i}={\bm\Sigma}_{r_i}^{1/2}{\bm z}_i$, we have $\|{\bm x}_i-{\bm\mu}_{r_i}\|\le \lambda_{\max}^{1/2}\|{\bm z}_i\|$, leading to $\|{\bm x}_i-{\bm\mu}_{r_i}\|^2\le \lambda_{\max} \|{\bm z}_i\|^2$, implying that ${\rm Pr}(\|{\bm x}_i-{\bm\mu}_{r_i}\|^2\le t)\le {\rm Pr}(\lambda_{\max}\chi_q^2\le t)$ for any $t>0$. We next obtain two inequalities similar to~\eqref{eq:upper probability kmeans spherical d>1} and~\eqref{eq:minimum difference inequalities for the initial centroids by the algorithm} using $\lambda_{\max}$ to replace $\sigma^2$, respectively. Then, we draw the conclusion for the case when $p\ge 2$. The proof for the case when $p=1$ is similar. By the combination of the two cases, we draw the conclusion. \qed

{\bf Proof of Corollary~\ref{cor:clustering consistency of kmeans QDA, p fixed}.} The conclusion can be directly implied by Theorem~\ref{thm:clustering consistency of kmeans QDA}. \qed

\end{document}